\documentclass[10pt]{article}
\usepackage{amsthm}
\usepackage{amssymb}
\usepackage{verbatim}
\usepackage{mathtools}
\usepackage{float}
\usepackage{bbm}
\usepackage{enumitem}   
\usepackage{lscape} 
\usepackage[utf8]{inputenc}
\usepackage[english]{babel}
\usepackage{biblatex}
\title{Bibliography management: \texttt{biblatex} package}
\author{Share\LaTeX}
\date{ }

\begin{document}

\begin{center}
  {\bf A Comparative Approach to Explainable Artificial Intelligence (XAI) Methods in Application to High-Dimensional Electronic Health Records: Examining the Usability of XAI} \\

  Jamie A. Duell 
\end{center}

\section{Introduction} \label{XAI} 

Machine Learning approaches are ubiquitous across many domains of application, altering the way that we approach real-world tasks, firstly, we acknowledge the two common forms of Machine Learning model designs, these are \emph{Black-box} and \emph{Glass-box}.

\begin{enumerate}[label=(\roman*)]

\item \emph{Black-box} models are a commonality for most ML algorithms, as they contain a sense of dissonance due to the unseen nature of mathematical model computation and impact on each input combined with a lack of understanding from domain experts, providing ambiguity in most cases to obtain the given output, these models often provide the best performance in trade for explainablility, as they do not provide an explanation as to why a solution is provided. Black-box models see the application of model-agnostic methods provided to interpret such relationships, there are various model-agnostic methods that are applied to provide an interpretation of machine learning methods extending the model to initialize a form of transparency regarding output, providing an explainable solution to a model that is otherwise not understood due to the black-box nature \cite{Gonzalez}, \cite{Rudin}. 

\item \emph{Glass-box} models imply that all necessary information is readily available, the trade-off for such method is often the models performance for an explainable solution, the term glass-box can often be thought of as synonymous with the term XAI. Glass-box methods have seen recent implementation with the likes of models such as the Explainable Boosting Machine (EBM) and DeepLIFT. EBM is a glass-box model introduced by Microsoft in 2019 \cite{nori2019interpretml}, model application provided in section \ref{baselinemodel}. The architecture of DeepLIFT can be used in junction with model-agnostic methods such as SHAP an explanatory method, this can be seen in the DeepSHAP implementation \cite{Chen}, this concatenation of both Shapley values and DeepLIFT is out of scope for the implementation of this paper.

\end{enumerate}

The aim of this paper is to apply Explainable Artificial Intelligence (XAI) methods to demonstrate the usability of an explainable tertiary layer appending black-box models. XAI aims to define the feature importance regarding a models prediction indicative of contribution. Therefore, the use XAI allows us to extract interpretable knowledge from high-dimensional data-driven questions to inform domain experts of trends that can be identified through the categorical and numeric features that can be difficult for a human expert due to data quantity, this intending to help inform decisions and place as a assistant - not a substitute to a human expert, forming a tertiary layer of knowledge, where we have: the patient data, the patient data observations from a domain expert; and an explainable solution to act as the aforementioned subsidiary tertiary layer. 

Notably, explainable solutions can also be regarded as interpretable, let us differentiate \emph{interpretable} and \emph{explainable}, in some cases these terms are used interchangeably, but for the purpose of clarity we define

\begin{enumerate}[label=(\roman*)]
\item \emph{Interpretable models} provide a visual representation of the data that determines the model findings, allowing the user to gauge an understanding of the output, this giving room to user interpretation as there's minimal clarity to reasoning, allowing a flexible view on  the results, this can differ between individuals.

\item \emph{Explainable models} are models that provide direct description as to the results and relationships that support the model prediction, as such simplifying and communicating a complex domain, introducing clarity and transparency, though human interpretation is not redundant as the provided explanations are model dependent and can see different results giving room for a human-expert. 
\end{enumerate}

This work will present state-of-the-art models that can be defined as explainable AI solutions, covering the usability of model-agnostic Machine Learning extensions to black-box models and glass-box model integration to provide data interpretability for large-scale Electronic Health Records (EHR). 

The medical field for one has seen a pursuit of interest for the use of Machine Learning (ML) algorithms, we can see challenges arise when dealing with the circumstance of human-subjects, the ideology behind trusting a machine to tend towards the livelihood of a human poses an ethical conundrum – leaving trust as the basis from humans towards the machine capabilities with extrapolation from patients to external parties such as insurance companies who hold investment in procedural success \cite{Asan}, the aforementioned paper highlights that human trust is a key element for human-AI collaboration ranging from the medical field to vehicle automation \cite{Gold}, as we idealise the application of autonomous behaviours it does pose questions to human response and acceptance of the integration of AI ubiquitous across domains of study and practicality, with the intent to provide a better performing solution than a human-expert alone, but a commonality arises in algorithm application, to where the model can exhibit different behaviours in a non-deterministic manner compromising the trust factor to where black-box models become an adversary to trust, though machines do often outperform human experts - how can we assure the AI solution is trustworthy? The work of Tonekaboni et al. \cite{Tonekaboni} surrounds the question of 'what clinicians want?' It is identified that merely having a highly accurate Machine Learning model is not sufficient to be adopted by clinical staff, notably a single metric such as classification accuracy does not provide insight to how the solution was obtained or provide depth to the models effectiveness \cite{Kim}, interpretability is not heavily required when a problem has been well studied, this indicating a per-case dependency varying in necessity when approaching fields such as medicine, where there a requisition of clarity needed due to the fragile nature of data in the field, with the field being less familiar with DL implementations. There’s a subset of algorithms that are commonly used as interpretable models, examples being: Linear regression, logistic regression and a decision tree, these providing interpretability on a modular level, with the Naïve Bayes classifier and k-Nearest Neighbours also being considered as interpretable due to distance observations on local data points providing a form of observational interpretability, but are less reliable explainable solutions \cite{Christoph}. 

\subsection{XAI and Understanding Explainability} \label{XAI2} 

The recent field of study XAI, the term was first introduced in 2004 \cite{Lent} by Lent et al., a subset of AI that aims to provide intelligible explanations to the end user, the need for XAI as defined in \cite{Adadi} applies as milestone of mitigation towards model doubt, to adhere to ethical concern or regulatory considerations that need to be made within the domain, should there be bias or discriminatory results – but, should we gain trust trust, we as humans could tend towards reliance and complacency \cite{Mueller} as we become comfortable with such systems as opposed to a supplement to human-expert domain knowledge. Appending to the given concerns, in an attempt supplement notable negative contributors with AI application, there is likelihood of regulation application for AI systems from a legality standpoint regarding black-box model integration as identified by the work of R. Hoffman et al. \cite{Hoffman} where it is deemed a “right to explanation” maybe a necessary protocol and the user rights must be safe guarded, as such XAI provides a milestone towards the direction of machine-based reasoning.  

The work of \cite{Chromik} defines an explanation as a social process subject to two roles taking fruition and two or more participants, the roles consisting of 1) a subject explaining and 2)a subject receiving the explanation, given this, human subject participation can hold two forms of contribution with explanation experimentation as identified in \cite{Chromik}, where one case – is the receival of feedback, provided to determine the usability of the explanation ensuring clarification of a pragmatic understanding of XAI explanation; the opposing approach being feedforward, this intends for the human participant to provide a general explanation, that the algorithmic explanation is compared against providing a benchmark of comparison. The research of Miller \cite{Miller} pinpoints the term XAI as a subset of Human-Agent Interaction located at the intersection of Social Science, Human-Computer Interaction and Artificial intelligence, the paper a strong endeavour, linking XAI with social science, where there’s exploration of the different levels that categorise a potential explanation, these used to identify where in the pipeline we’re defining the posed question that derives our explanation – with regards to answering the question. The intent of this study, is to provide XAI generated explanations for the given predicted results of the patient case decision, producing a determinant as to why the case is predicted like-so for further domain-expert evaluation.
Referring to user interaction, we must obtain the correct mechanisms of approach to gain useful feedback,  one form of posing a question is providing confidence intervals that can be assessed through the feedback, this seen for a medical case-study, with the implementation seen in the works of \cite{Lamy}, where domain-experts have been assessed, by being presented cases to determine on a 5-value Likert scale the confidence of the decision made for a given explainable solution, Tonekaboni et al. \cite{Tonekaboni} identify three key metrics for evaluating XAI, these being

\begin{enumerate}

\item \emph{Domain Appropriate Representation}; identifying the coherence of representation with respect to a domain specific task, ideally nullifying redundancy in explanation, therefore, we can identify from the applied area what information would be valuable to present in explanation, in essence filtering the response to the user.

\item \emph{Potential Actionability}; idealistically inform clinical workflow via explanation, should the model itself present an account of basic trust principles, as factors such as patient similarity is not informative, usable, and can in times cause unnecessary cognitive overload, as such, XAI should place explanations that can lead to decisions with identification of factors that allow for follow up workflow.  

\item \emph{Consistency}; aligns with logical mapping of approximate 1-1 domain knowledge to model explanation, where changes in prediction can yield a representational change in explanation.

\end{enumerate}

It’s stated that there’s already awareness in the clinical field that there is no perfect prediction, but sole focus from this paper defines an acceptable explanation as one that falls in line with justification of their clinical decision making that align with medical practice, it is also noted that local explanations are highly valued, with a preference for patient individuality and feature importance with regards to the singular instance.
A method of approach for future work would be the introduction of a user-study in response to being provided explanations, this would ideally be an amalgamation of the aforementioned feed-forward and feed-backward methods, as the initial approach of feedforward will define a benchmark of explanatory comparisons based on input feature predictions to where we can gauge a human-expert opinion as a blind comparison, the human-expert will then see the given explanation to see if the explanations holds a form of usability and if anything is newly identified from the given explanations, the intent to provide explanations from each algorithm implementation to determine the most useful. The purpose behind such intent is that curiosity is noted as a key aspect of seeking an explanation \cite{Hoffman}, as, if the explanation is provided initially, it could skew the natural thought process of human reasoning and modify the answer post explanation. Therefore, to mitigate such risks, the intent in future work would be to first gain user explanation, as factors such as trust and curiosity can be measured through user study to gain insight on the XAI model applicability, as agreeing with the model is one step, but the key element of progress for XAI application is the acceptance of explanation from human-emotional response, this an identifiable gap for future work.
For the demonstrative purpose of this paper, we define the objectives of the study to be 1) To demonstrate the explainable visualization of XAI for high-dimensional medical data. 2) To contrast the different model-agnostic methods and the extraction of feature importance, providing commonalities and differentiation amongst each.  3) To determine pros and mitigate cons of the XAI model to provide aid to human-experts approaching high-dimensional data. Given these objectives, we introduce four XAI models; three being model-agnostic black-box extensions and one a glass-box method, with the ideology surrounding the demonstration of XAI as a supplementary tertiary layer for domain-expert interpretation, enabling the consideration of practical application in the medical field, the XAI methods are namely: Shapley Additive Explanations (SHAP), Local Interpretable Model-Agnostic Explanations (LIME), Scoped Rules (Anchors) and Explainable Boosting Machine (EBM).

XAI proposes a form of reasoning, this is often communicated in natural language and through the use of diagrammatic representation, identifying the route associative to the decision making process of the ML model, given that the representation of a models performance in traditional ML application does not provide complete reasoning on feature importance and the correlation between the model inputs and the model output..

\section{Method} \label{dev}

\subsection{Case Study and Data Pre-processing}
Though existing models are effective in providing a predictive accuracy that can be deemed as successful, there is a certain scepticism when it comes to interpreting data from any Machine Learning (ML) model, vulnerabilities can be projected in the field of medicine from a misinterpretation of results and findings, it’s critical when dealing with patient data that clarity is portrayed by the model alone to validate the given results so relationships between direct patient features and external associations such as regimen alterations, this issue can be approached by developing a transparent and idealistically transferable model, not a specific domain altered model, this defined as an Explainable Artificial Intelligence (XAI) solution, with the intent to communicate model correlations determining the route taken to obtain the results. Conclusively, XAI needs to be explored in the field of healthcare to enable better transparency giving clear reasoning-based performance.

The systematic intent behind the development of the model is to obtain useful information from a given domain. This environment space will contain features of large dimensionality to deduce a prognostication of trends, patterns and reasoning once the models have been applied, to extract valuable information to answer specific questions surrounding a given domain, this will be an aggregation of many feature inputs, to obtain useful information from a given research question.

Data for this paper used artificial data from the Simulacrum, a synthetic dataset developed by Health Data Insight CiC derived from anonymous cancer data provided by the National Cancer Registration and Analysis Service (NCRAS), which is part of Public Health England, this will be used to develop the XAI framework. The Simulacrum data set consists of 1,322,100 synthetic patients allowing for model development for researchers whilst maintaining clear patient confidentiality, reflecting a high degree of accuracy reflective of properties found in the NCRAS data set, allowing for transferable developed models from synthetic data sets to real-world data sets.

The adoption of the Simulacrum data set enables the capabilities for ML model development on big data, with initial steps of the project involving data pre-processing to allow for the data to use ML effectively, the initial phase of development prioritised the creation of a master table concatenating both Cancer Registration (AV) and Systemic Anti-Cancer Therapy (SACT) tables with a total of 63 columns, the available tables with the associated number of columns is illustrated in table \ref{table:1}. 
\begin{table}[H]
\centering
 \begin{tabular}{|c | c|} 
 \hline
 \textbf{Table Name} & \textbf{No. Associated Columns} \\  
 \hline\hline
 sim av patient & 6 \\ 
 \hline
 sim av tumour & 16 \\ 
 \hline
 sim sact patient & 2 \\ 
 \hline
 sim sact tumour & 5 \\ 
 \hline
 sim sact cycle & 6 \\ 
 \hline
 sim sact drug detail & 10 \\ 
 \hline
 sim sact regimen & 10 \\ 
 \hline
 sim sact outcome & 8 \\
 \hline
 \end{tabular} 
\caption{An overview of the available Simulacrum data set tables with the numeric value of associated columns}
\label{table:1}
\end{table}

Firstly, we must deduce a set of different medical questions to apply the base ML methods and XAI framework, therefore, we identified a set of four supervised classification problems, two with a focus on all given patients, and two reduced to a specific cohort of patients, these namely: 
\begin{itemize}
\item \emph{Predicting the likelihood of reducing any patients drug dose.}
\item \emph{Predicting the likelihood of mortality in any given patient.} 
\item \emph{Predicting the likelihood of reducing a lung-cancer patients drug dose.}
\item \emph{Predicting the likelihood of mortality for lung-cancer patients.} 
\end{itemize}

Following this, the intervention of data set manipulation allows for the construction of a machine teachable model, this is antecedent to the application of ML methods themselves, in the sense of morphing the existing data with methods such as encoding, this step is to provide data that removes null values or hard to teach unexplanatory patterns, instances in this case could include logical inconsistencies such as having a patients weight or height that is illogical of being, and cases where undergoing a regimen after death, or vital status being claimed as alive at such time, this being reflective of real-world noisy data allowing for beneficial practice of technique implementation, allowing for a viable model demonstrator. Following this, the data set is then balanced to the lower bounds bias for each binary classification output, covering lung cancer (LC) patients for both mod dose reduction (MD) and mortality prediction (DA), and all patients. 

\begin{table}[H]
\begin{center}
 \begin{tabular}{| c | c c c c|} 
 \hline
 \textbf{Bounds} & \textbf{LC-DA} & \textbf{LC-MD} & \textbf{MD} & \textbf{DA} \\
 \hline\hline
 Lower & 25456 & 21567 & 181626 & 384375 \\ 
 \hline
 Higher & 95944 & 27752 & 1034230 & 830434 \\ 
 \hline
 Updated Higher & 24000 & N/A & 180000 & 380000 \\ 
 \hline
\end{tabular}
\caption{An overview of the updated output balance for each question}
\end{center}
\label{table:1}
\end{table}

We compose the given data set in a tabular format where each row corresponds to a singular patient visit, this singular input will be comprised of a set of features that will be applied to each question posed without selecting features based on question bias, therefore, the set of features fed into the model, question dependant should extract a different level of feature importance given the direction of the question. Ideally, feature dependency will alternate between each question focus, as such we can assume association has been appropriately made regarding domain specificities. Therefore, we define a set of input features from the Simulacrum data set, these identified in table 3 with the name in the original data set to the altered name for output clarity, supported with a description of the features, the feature name '\emph{NEW VITAL STATUS DESC}' is used as the supervised output for mortality, whereas '\emph{REGIMEN MOD DOSE REDUCTION}' is used for the supervised output for the reduction of drug dosage, these are not inversely applied as input features to the opposing models, to keep inputs to the model synchronous. 

\subsection{Baseline Model Implementation} \label{baselinemodel}
\begin{figure}[H]
\centering
\includegraphics[scale=0.35]{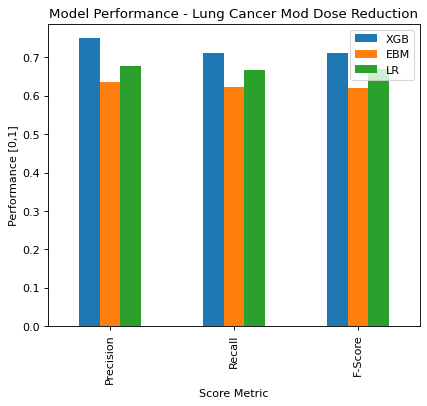}
\includegraphics[scale=0.35]{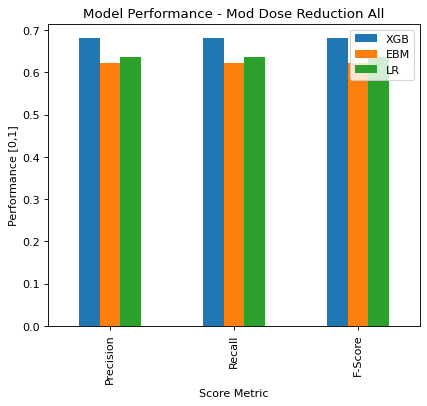}
\includegraphics[scale=0.35]{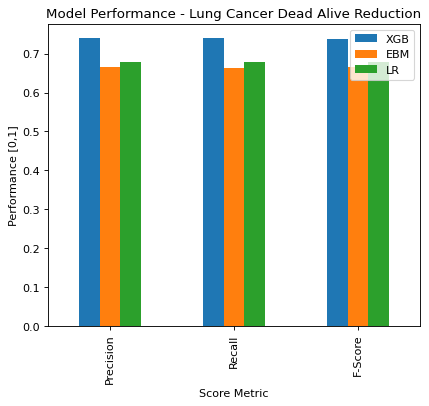}
\includegraphics[scale=0.35]{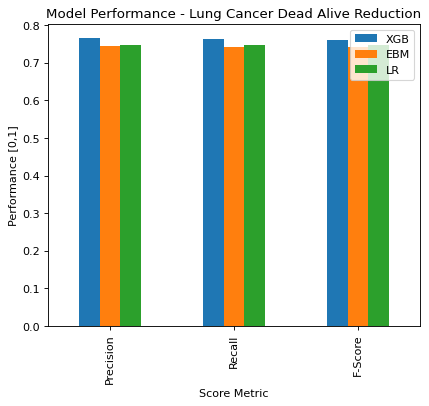}
\caption{Comparison of algorithm performance across each available data set}
\label{fig:comparison}
\end{figure}

To obtain results from model-agnostic methods we must first execute black-box algorithms on the given data sets, the two decided algorithms which will inherit an explanation are Logistic Regression and XGBoost, this to which we can compare against a glass-box method an Explainable Boosting Machine. Therefore, provided in the summative bar charts is a comparative approach identifying base performance metrics being precision, recall and accuracy of each applied baseline model, these shown in figure \ref{fig:comparison}.
\begin{table}[H]
\begin{center}
 \begin{tabular}{|c | c c c c|} 
 \hline
 \textbf{Dataset} &      & \textbf{Precision} (\%) & \textbf{Recall}(\%)  & \textbf{Accuracy}(\%)  \\ [0.5ex] 
 \hline\hline
       & \emph{Logistic Regression} & 75 & 75 & 75 \\ 
 \textbf{DA} & \emph{XGBoost} & 76 & 76 & 76 \\ 
       & \emph{EBM} & 74 & 74 & 74 \\ 
 \hline
        & \emph{Logistic Regression} & 68 & 68 & 68 \\ 
 \textbf{LC-DA} & \emph{XGBoost} & 78 & 78 & 78 \\ 
       & \emph{EBM} & 67 & 67 & 67 \\ 
 \hline
       & \emph{Logistic Regression} & 64 & 64 & 64 \\ 
 \textbf{MD} & \emph{XGBoost} & 68 & 68 & 68 \\ 
       & \emph{EBM} & 62 & 62 & 62 \\ 
 \hline
       & \emph{Logistic Regression} & 68 & 67 & 67 \\ 
 \textbf{LC-MD} & \emph{XGBoost} & 75 & 71 & 71 \\ 
       & \emph{EBM} & 64 & 62 & 62 \\ 
 \hline
\end{tabular}
\caption{Summary table of baseline performances for logistic regression, XGBoost an EBM tested on each dataset, where DA = All patients for mortality prediction, LC-DA = Lung Cancer patients for morality prediction, MD = Reduction of drug dosage for all patients and LC-MD = Lung Cancer patients for the reduction of mod dosage }
\end{center}
\label{table:1}
\end{table}

From this comparison, we can deduce that the XGBoost algorithm is the best performing  across all the given datasets, this can be applied as the baseline algorithm for the extension of model-agnositc solutions. 

\subsection{Model-Agnostic Techniques and Application} \label{model technique and app}
\begin{figure}[H]
\centering
\includegraphics[scale=0.5]{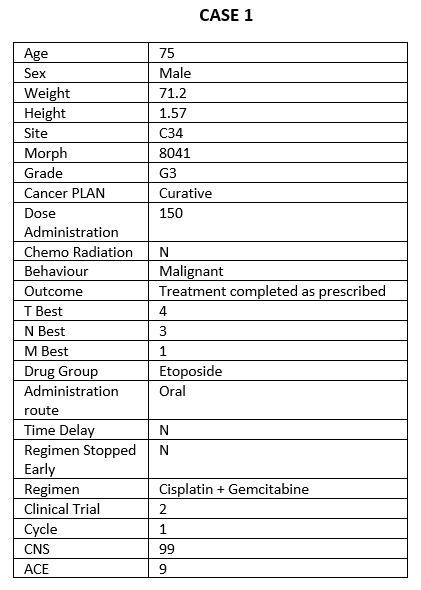}
\caption{Case study for explanation demonstrations for predicted case: Deceased | Actual case: Deceased}
\label{fig:case}
\end{figure}

For the formulation of XAI techniques on the posed questions, we apply three different model-agnostic techniques to obtain the decisive input features for each question further evaluated in section \ref{analysisofmodel}, for the introduction of the model-agnostic models - SHAP, LIME and Scoped Rules (Anchors), the visualization examples will be provided from a single case-study from the LC-DA dataset and apply each method with the supported explanations as a demonstrator, the case defined in figure \ref{fig:case}, with supporting explanations in sections \ref{SHAP}, \ref{LIME} and \ref{SR}. 

\subsubsection{Shapley Additive Explanations} \label{SHAP}

Shapley Additive exPlanations (SHAP) is a model-agnostic framework introduced by Lundberg et al. \cite{Lundberg} that is intended to explain ML algorithms through visualization, introduced 2 years posterior to LIME, with a similar intent as LIME aiming to highlight feature relevancy, re-instantiating how there's an common goal of gaining user trust by obtaining an explanation.

The intent of SHAP is to explain the given prediction through calculating the feature importance for the ML prediction. The use of Shapley values directs the model on how to fairly distribute the feature or a group of features importance which we determine as players in game theory, therefore, importance is based on the player investment, the implementation of Shapley value sees the incorporation of a characteristic function game G, for a number of players {N = 0,1,2,..\(n\)}, ensuring that coalitions \begin{math} G \subseteq N \end{math}, where the shapley value calculates average contribution over a permutation of players. Notably, additive explanations can be defined as a linear function of binary inputs, represented

 \[ g(z^\prime) = \phi_0 + \sum_{k=1}^{M} \phi{_k}z{^\prime_k}  \]

\begin{figure}[H]
\centering
\includegraphics[scale=0.4]{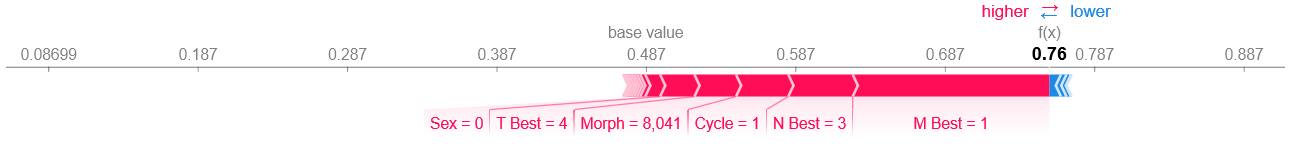}
\caption{SHAP providing an explanation for a patient, with the width of each descriptive block and colour are indicative of the shift in probability to a given case. In this case Red indicating a shift towards death which is countered by a small adjustment of blue, providing an overall function of (x) providing the likelihood of the patient being deceased. }
\label{fig:shaplocal}
\end{figure}

 Where we identify g as the given explanation for an original prediction f(x) where \(z{^\prime} \subseteq\{0, 1\}^M\), to where M is the number of simplified inputs, explanatory models use simplified feature inputs due to the complexity of the original input data, finally we have \(\phi_k\), this is feature attribution for feature \(k\), we then obtain \(g(z^\prime)\) the sum of all feature contributions for the linear model g which is fit for the simplified application of \(z'\). The localised contribution of features on the identified case produces the output displayed in \ref{fig:shaplocal}. 

\subsubsection{Local Interpretable Model-agnostic Explanations} \label{LIME}

Regarding extensions to ML models to provide an explainable model, Local Interpretable Model-agnostic Explanations (LIME) is an example of such model agnostic method that was proposed by Ribeiro et al. \cite{Ribeiro2} that explains a choice indicating feature importance towards a decision, providing somewhat of an explanation towards the solution, LIME can be applied to any ML model, with the intent to understand the model based on how the prediction changes as it passes through the model to obtain the output based on the input data samples, notably, when LIME is used, it’s still required that the end-user must interpret the output correctly.

Surrogate models have the goal of providing an approximation for the predictions of a black-box machine learning model, the explanation model minimizes the loss and measures whether or not the explanation is close to the prediction of the original model, it's notable that the model only optimizes the loss \cite{Christoph}. LIME can be applied to any type of given data, the model of LIME namely focuses on the local interpretablility of a model, this by accessing a single input feature a fitting a line of linearity using a regularization constraint to the linear regression model, we try to find \(\phi\) by minimising loss, where we measure the faithfulness of the explanation \(g\) to the original \(f(x)\) for a local prediction, we can view this as \(L(f, g, \pi{_x})\) applying the regularization parameters \(\Omega(g)\) where L denotes the square loss function to which we try minimize, therefore, we define the explanation for a local point \(x\) using LIME as 

\[E = arg \ min \ L(f, g, \pi{_x}) + \Omega(g) \]

\begin{figure}[H]
\centering
\includegraphics[scale=0.4]{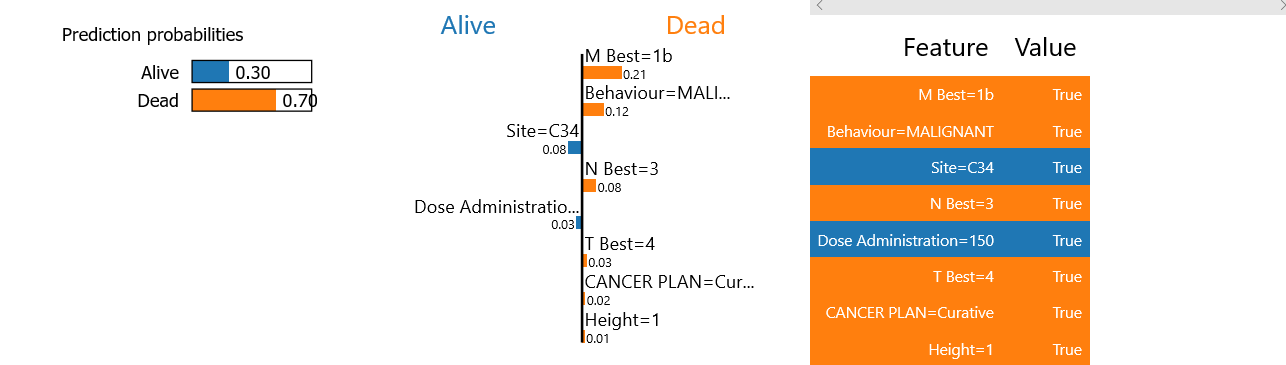}
\caption{LIME providing an explanation given probability of Alive / Deceased cases providing in the top left, supported by each feature value and supporting importance value with the corresponding class to which each feature is weighted towards.}
\label{fig:LIME}
\end{figure}

In practice, we can extract local explanations using LIME determining the considered most important features for the local linear model, the left of Figure \ref{fig:LIME} showing the probability for towards each classification, with the right displaying each feature importance towards the outputs, in the given example we see M Best and Behaviour decreasing the likelihood of survival, with the largest value of importance, represented by the block length and contributing value 0.21, 0.12, contrasting to that of the feature that shift probability of living, in turn concluding to a large death probability, notably, the initial 8 priority features are represented, it is possible to show the entirety of feature contribution, the figure intent demonstrates the communicative capabilities of XAI visualization. 

\subsubsection{Scoped Rules} \label{SR}

Scoped rules (Anchors) introduced in 2018 by Ribeiro et al. \cite{Ribeiro} is used to explain individual predictions of a classification model by prioritising a search for a decision rule to increase the prediction likelihood. The intent behind anchors to integrate, much like LIME in section \ref{LIME} a perturbation-based strategy is used to create an explanatory output of black-box machine learning models. Anchors follow the idea of factual explanations, identifying cases where certain anchors given a condition fit to obtain the an output. 

Differentiation from LIME is evident in the explanation structure, providing a rule-based format of communication with base IF-THEN rule, these being the anchors. The introduction to such extension of LIME provides a better localized understanding. An Anchor can be defined as followed:

\[\mathbb{E}_{D_x}{}_{(z|A)} [\mathbbm{1}_{f(x)=f(z)}] \geq \tau, A(x)=1 \]  
Where predicate \(A\) is an anchor if the expected evaluation of the neighbours of instance \(x\) of the distribution \(D\) matching \(A\) is greater-than or equal-to the precision boundary set on \(\tau\), therefore, only rules that achieve this local fidelity are viable and then will be considered an anchor. 

An example of Anchor usability demonstrated in \cite{Ribeiro} shows the prevalence of Anchors for natural language processing application, with local instance of a word such as "bad" or "not" have negative connotation, though not reflecting upon the full statements sentiment, an example being "This movie is not bad" this could be misinterpreted to have negative sentiment due to isolated words with an unseen region, this could put salience using weights on the word "not" and correlate the statement to a negative sentiment,  whereas, the introduction of anchors attempts to associate word relations through coverage of segmented phrases such as {"not", "bad"} having a positive connotation explanation and {"not", "good"} having a negative connotation explanation, which could otherwise be misinterpreted, especially if the instances are presented to the user with no structure, we will proceed to identify the explanation on a tabular data set, as presented by the given case.

\begin{figure}[H]
\centering
\includegraphics[scale=0.5]{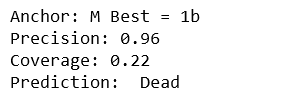}
\caption{Anchors gives a conditional conjunction of cases these being Anchors which identify given certain elements are true, these Anchors play importance on the outcome. In this case there is a single Anchor, but this can be a string of 'AND' conditions that would be prevalent in some other cases.}
\label{fig:AnchorLocal}
\end{figure}

\subsubsection{Analysis of model-agnostic Application}\label{analysisofmodel}

\begin{figure}[H]
\centering
\includegraphics[scale=0.5]{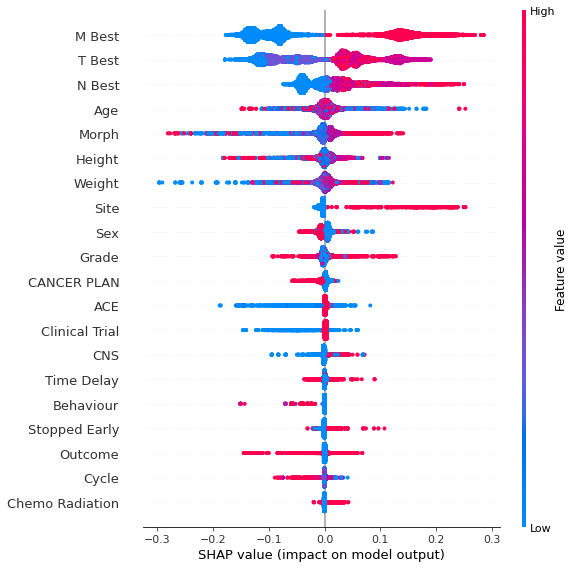}
\caption{SHAP global explanation, the x-axis provides a weighting with positive SHAP values shifting towards deceased and negative SHAP values shifting towards death, with the width of the given data point scattering to the level of importance it has for each instance, therefore ‘0.0’ indicative of minimal impact. The feature value is also represented from low to high based on colour as shown on the right vertical bar, this for each feature over the entire dataset.}
\label{fig:shap}
\end{figure}

\begin{figure}[H]
\centering
\includegraphics[scale=0.4]{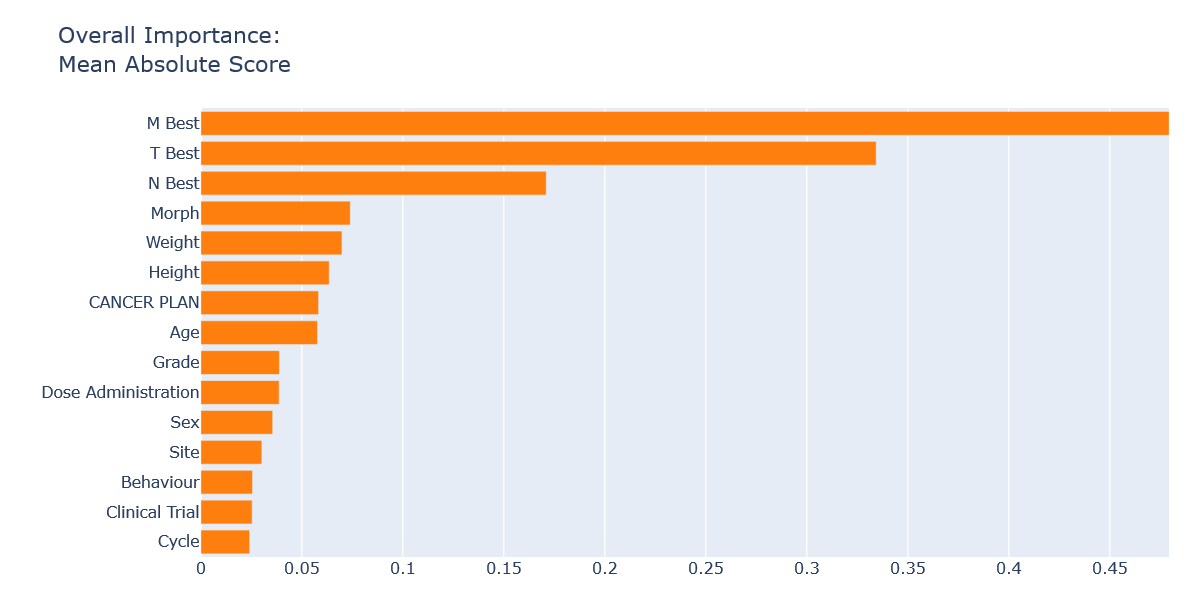}
\caption{EBM global explanation, the x-axis shows the mean absolute importance for the given features, indicative of the impact of each feature over the entire LC-DA dataset.}
\label{fig:EBM}
\end{figure}

We apply these listed three model-agnostic solutions, to the given four questions to provide a comparative approach for feature importance each a binary classification problem. Through the use of SHAP and EBM, we can introduce the extraction of global feature importance, the given example was trained using XGBoost on Lung Cancer patients in the binary classification ["Dead", "Alive"] introduced in section \ref{dev}, the SHAP model in Figure \ref{fig:shap} denoting negative weighting towards output solution on the left or positive on the right, of 0 on the \(x\) axis, shifts towards an output, with a colour gradient representation of the feature value, with the next image in Figure \ref{fig:EBM} demonstrating the absolute value of feature importance through the use of EBM to provide a comparison on feature ordering and representation of global importance.

Firstly, we have LIME this isolating an individual patient instance as a local linear fit. Through the LIME framework we can provide a view of feature importance for a single instance as interpreted by the LIME model, this then explained the end user,   

Secondly, we then apply SHAP, where we view each instance of both the feature and feature value of a singular record  as the 'player' to determine the players contribution to the obtained output, therefore, in our context we view each tabular category e.g "Site=C50", "Age=70"etc. as the players,  we can then obtain the contribution towards one of the two output classes, in this case being either ['Y', 'N'].

\begin{figure}[H]
\centering
\includegraphics[scale=0.35]{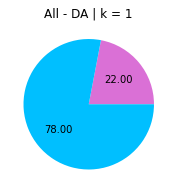}
\includegraphics[scale=0.35]{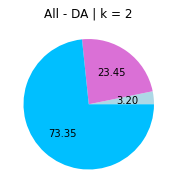}
\includegraphics[scale=0.35]{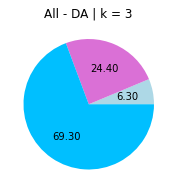}

\includegraphics[scale=0.35]{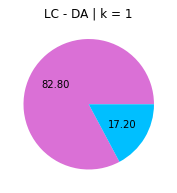}
\includegraphics[scale=0.35]{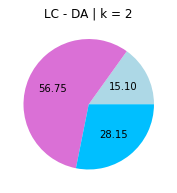}
\includegraphics[scale=0.35]{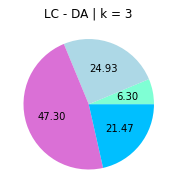}

\includegraphics[scale=.5]{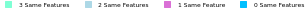}
\caption{Comparing the shared features across both DA data sets using SHAP and LIME}
\label{fig:comparisonnew2}
\end{figure}

\begin{figure}[H]
\centering
\includegraphics[scale=0.35]{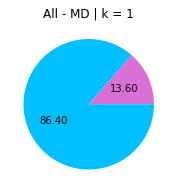}
\includegraphics[scale=0.35]{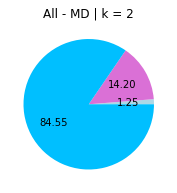}
\includegraphics[scale=0.35]{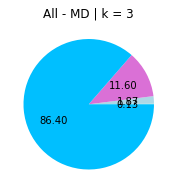}

\includegraphics[scale=0.35]{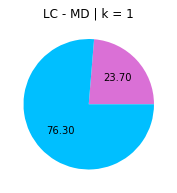}
\includegraphics[scale=0.35]{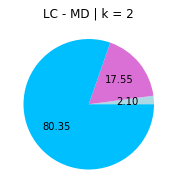}
\includegraphics[scale=0.35]{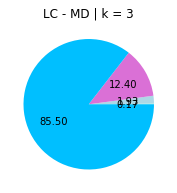}

\includegraphics[scale=0.5]{images/legend2.png}
\caption{Comparing the shared features across both MD data sets using SHAP and LIME}
\label{fig:comparisonnew}
\end{figure}

We can view the shared features by absolute value irrespective of positive or negative connotation towards to the model result, where they are contributing in order of importance such that the named data set e.g. MD, for models SHAP and LIME share a named feature \(F\) concluding that \(SHAP(F_k)\) == \(LIME(F_k)\) where \(k\) is the selected feature index, we can conclude this for multiple features by returning shared features up to \(k\), for each data set we allow for where k = 1, k = 2 and k = 3, these demonstrated for each data set independently shown in Figures \ref{fig:comparisonnew2} and \ref{fig:comparisonnew}. 

\begin{figure}[H]
\centering
\includegraphics[scale=0.35]{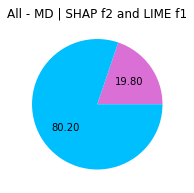}
\includegraphics[scale=0.35]{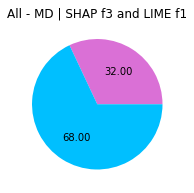}

\includegraphics[scale=.5]{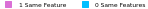}
\caption{Comparing the shared features on the MD data set using SHAPs second and third feature list and LIMEs first}
\label{fig:diffeat}
\end{figure}

\begin{figure}[H]
\centering
\includegraphics[scale=0.3]{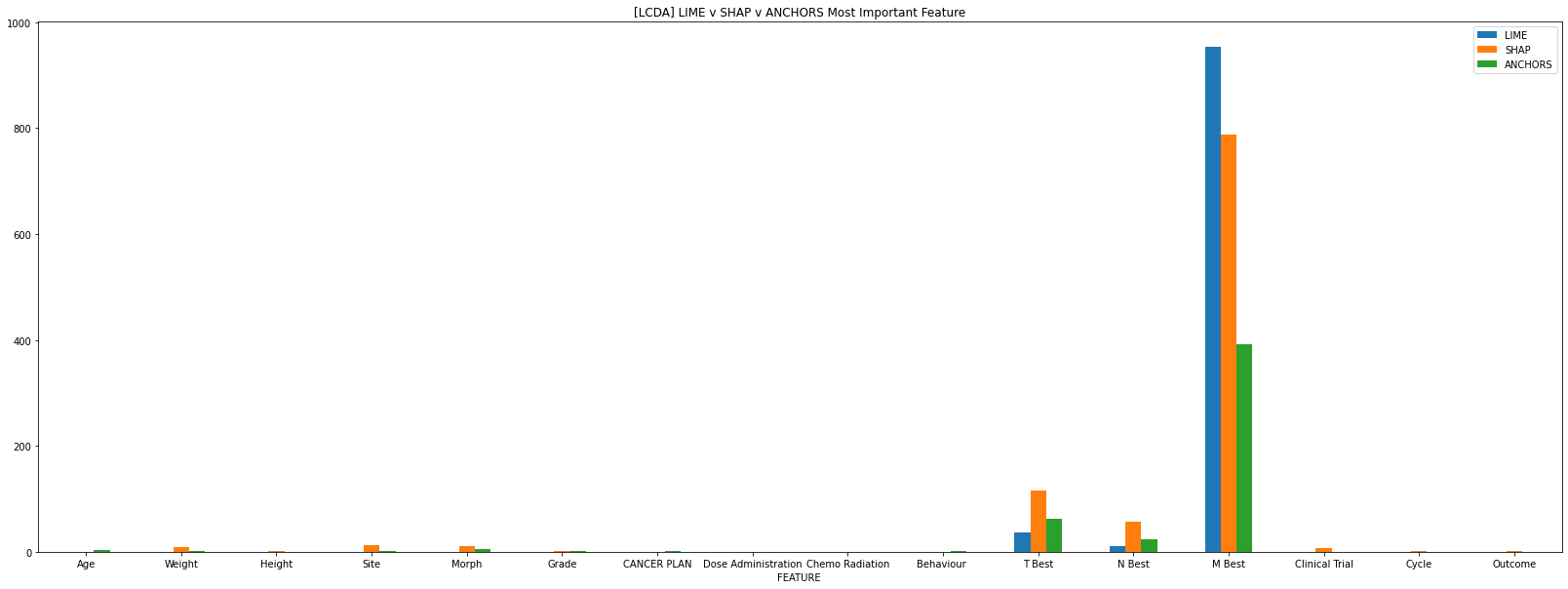}
\includegraphics[scale=0.3]{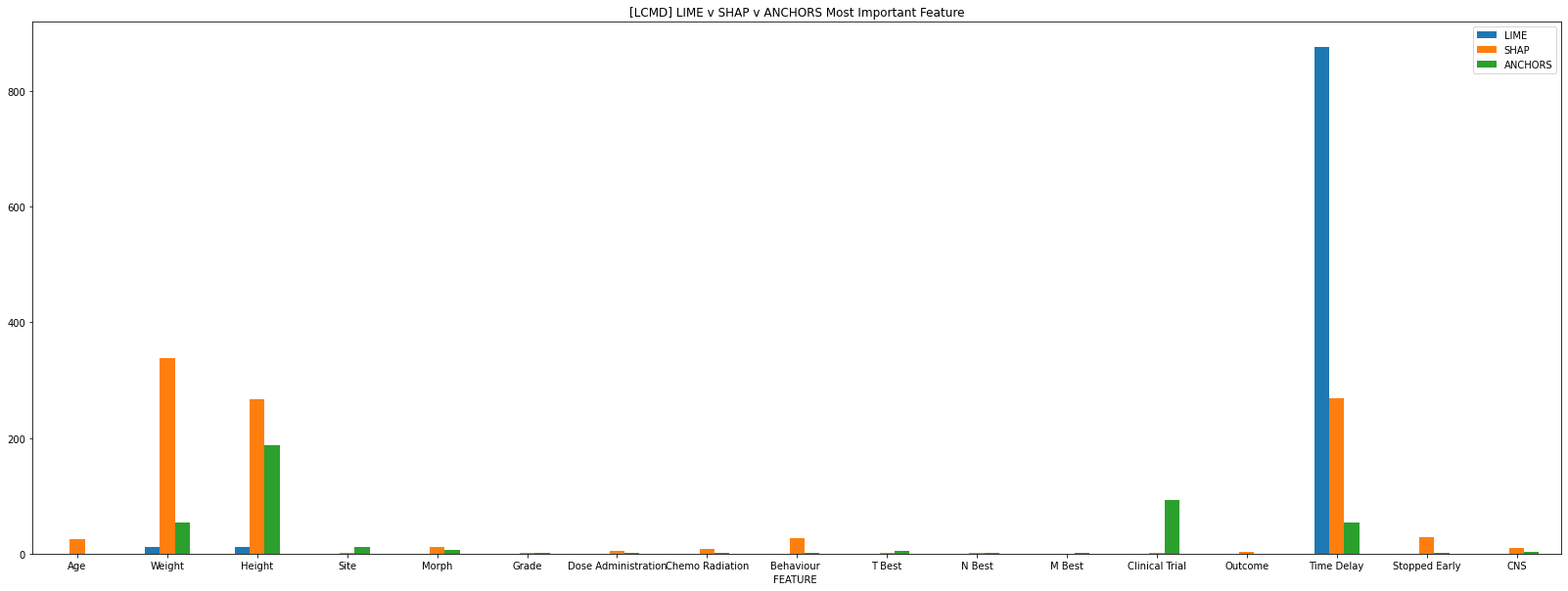}
\caption{Most important feature returned or the first anchor (scoped rules) for the first 1000 instances on the test data set for both LC-DA and LC-MD data sets}
\label{fig:comparenumbers}
\end{figure}

From these shared features we can see inconsistencies amongst the shared features, though given this fact, inverse comparisons where e.g. \(SHAP(F_1)\) == \(LIME(F_2)\) there are a high majority of features, as priority on the most important feature, as shown in the first 1000 iterations in figure \ref{fig:comparenumbers}, it graphs that the most important feature is the same feature for a high percentile of the designated population, it would seem that the most important feature in some cases, aren't inline with one another, to draw such comparison we view the MD data set as an example where the shared features for k = 1 provides the smallest percentage, therefore we compare \(SHAP(F_2)\) and \(SHAP(F_3)\) against \(LIME(F_1)\) providing evidence of differential in priority order, returning the shared features shown in Figure \ref{fig:diffeat}. 

Given such information, we can also determine priority features from the data sets, though they may not be shared for each instance, it's optimal that the identified important features are relational to a certain extent for validation supporting the consistency of knowledge representation. Therefore, we provide a comparison across the four data sets, where we take the first 1000 instances from the test data set to extract the most important feature where \(k = 1\) for each LIME and SHAP as well as the first Scoped Rules anchor, where we determine the most influential factors towards a patient either ["Dead", "Alive"] or have reduced drug modification for the current regimen ['Y', 'N']. 

Evidently, commonalities arise across the shared most important feature for each XAI application, only matched ordering for features priority is less common for each model, this differentiating in balance across the data sets, where we see SHAP holding priority on weight for the LC-MD data set conversely to LIME having Time Delay as the greatest priority, where we see a more balanced dissemination across the three features for the SHAP model, notably, LIME shares percentiles of the same importance regarding features, we can see that Scoped Rules also determines shared anchors across the same set, this displayed in Figure \ref{fig:comparenumbers}. 

\section{Future Work and Conclusion}

Determining feature priority from a local interpretation provides a sense of clarity for non-domain experts with a base clarity of feature contribution to a given case. As appointed in section \ref{model technique and app} we derive a slight aberration in the feature attribution priority order amongst the model-agnostic solutions, though still sharing a level of importance across three features. For future work, to support the discernible necessity of human-expert trust, we aim to carry out a user study to determine effectiveness of the given explanations, the direct extraction of \(k=3\) feature importance comparison provides discomfort in reliability as it poses questions towards the validity of explanation, as such it was then shown a majority of features are shared across those 3 important features, only not in junction for all use cases, this providing support of clear commonality, given the data is synthetic, the provided explanations are that of demonstrative visualization and explanation usability, that from future work can determine validity. 

The ability to extract explanations of black-box model predictions is essential for high-risk applications of machine learning; medical application being one example, from such information of patterns across features and different explanation models we can determine data set and supporting predictions fairness. Explanation locality allows for new instances to be communicated to the domain-expert with reason. This tertiary layer of knowledge could improve the rate of case deduction and support human-expert reasoning.  

\printbibliography

\section{Appendix}
\begin{landscape}
\begin{table}[H]
 \begin{tabular}{|c | c | c |} 
 \hline
 \textbf{Feature Name in Data set} & \textbf{Feature Name Transformed} & \textbf{Feature Description} \\
 \hline\hline
 AGE & Age & Age of the patient at the time of instance \\ 
 \hline
 SEX DESC y & Sex & Sex of the patient  \\ 
 \hline
  WEIGHT AT START OF REGIMEN & Weight & Weight in kg. at the start of the SACT regimen  \\ 
 \hline
  HEIGHT AT START OF REGIMEN & Height & Height in metres at the start of the SACT regimen  \\ 
 \hline
  SITE ICD10 O2 DESC & Site & The given ICD-10 Code  \\ 
 \hline
  MORPH ICD10 O2 & Morph & Morphology ICD-1O code \\ 
 \hline
  GRADE & Grade & Indicative of Tumor grade \\ 
 \hline
  CANCER CARE PLAN INTENT & CANCER PLAN & The cancer plan  \\ 
 \hline
  ACTUAL DOSE PER ADMINISTRATION & Dose Administration & Dose in mg for each administration in the SACT cycle  \\ 
 \hline
  CHEMO RADIATION & Chemo Radiation & Whether or not a patient is undergoing radiotherapy  \\ 
 \hline
  BEHAVIOUR ICD10 O2 DESC & Behaviour & Behaviour description  \\ 
 \hline
  REGIMEN OUTCOME SUMMARY DESC & Outcome & A description of regimen outcome \\
 \hline
  NEW VITAL STATUS DESC & STATUS & Whether the patient is Dead or Alive  \\ 
 \hline
  T BEST & T Best & The size and extent of the tumor \\ 
 \hline
  N BEST & N Best & No. of nearby lymph nodes that have cancer  \\ 
 \hline
  M BEST & M Best & Whether the cancer has spread from the primary tumor  \\ 
 \hline
  DRUG GROUP & Drug Group & The drug group name \\ 
 \hline
  ADMINISTRATION ROUTE DESC & Administration route & Method of delivery for each cycle administration  \\ 
 \hline
   REGIMEN MOD TIME DELAY & Time Delay & Time delay between administration  \\ 
 \hline
   REGIMEN MOD STOPPED EARLY & Stopped Early & Regimen stopped early \\ 
 \hline
   MAPPED REGIMEN & Regimen & The mapped regimen name \\ 
 \hline
   CLINICAL TRIAL & Clinical Trial &  Patient is in an active Systemic Anti Cancer therapy trial  \\ 
 \hline
   CYCLE NUMBER & Cycle & The current cycle  \\ 
 \hline
   CNS & CNS & CNS \\ 
    \hline
   REGIMEN MOD DOSE REDUCTION & Reduction & Identifies the reduction of the drug dose during regimen  \\ 
 \hline
   ACE27 & ACE & Adult Co-morbidity Evaluation  \\
 \hline
\end{tabular}
\caption{Features with given description}
\label{table:1}
\end{table}
\end{landscape}
\end{document}